\documentclass[10pt,twocolumn,letterpaper]{article}

\usepackage[pagenumbers]{iccv} 

\usepackage{multirow}
\usepackage{makecell}
\usepackage{color, colortbl}
\definecolor{lgray}{rgb}{0.9,0.9,0.9}

\definecolor{iccvblue}{rgb}{0.21,0.49,0.74}
\usepackage[pagebackref,breaklinks,colorlinks,allcolors=iccvblue]{hyperref}

\def\paperID{11816} 
\def\confName{ICCV}
\def\confYear{2025}

\title{EMF: Event Meta Formers for Event-based Real-time Traffic Object Detection}

\author{
Muhammad Ahmed Ullah Khan\thanks{Authors contributed equally to this work.},
Abdul Hannan Khan\footnotemark[1],
Andreas Dengel\\
Department of Computer Science, RPTU Kaiserslautern-Landau,\\
German Research Center for Artificial Intelligence (DFKI GmbH),\\
67663 Kaiserslautern, Germany\\
{\tt\small Corresponding Author: hannan.khan@dfki.de}
}

\begin{document}
\maketitle
\begin{abstract}

Event cameras have higher temporal resolution, and require less storage and bandwidth compared to traditional RGB cameras. However, due to relatively lagging performance of event-based approaches, event cameras have not yet replace traditional cameras in performance-critical applications like autonomous driving. Recent approaches in event-based object detection try to bridge this gap by employing computationally expensive transformer-based solutions. However, due to their resource-intensive components, these solutions fail to exploit the sparsity and higher temporal resolution of event cameras efficiently. Moreover, these solutions are adopted from the vision domain, lacking specificity to the event cameras. In this work, we explore efficient and performant alternatives to recurrent vision transformer models and propose a novel event-based object detection backbone. The proposed backbone employs a novel Event Progression Extractor module, tailored specifically for event data, and uses Metaformer concept with convolution-based efficient components. We evaluate the resultant model on well-established traffic object detection benchmarks and conduct cross-dataset evaluation to test its ability to generalize. The proposed model outperforms the state-of-the-art on Prophesee Gen1 dataset by $1.6\text{ }mAP$ while reducing inference time by $14\%$. Our proposed \textit{EMF} becomes the fastest DNN-based architecture in the domain by outperforming most efficient event-based object detectors. Moreover, the proposed model shows better ability to generalize to unseen data and scales better with the abundance of data.

\end{abstract}    
\section{Introduction}
\label{sec:introduction}

Camera-based perception for autonomous driving is one of the major applications of computer vision. Recent advancements in the field have elevated the performance of such perception systems to a new level. ViT \cite{dosovitskiy2020image} and advanced convolution-based architectures like, ConvNeXts \cite{liu2022convnet}
have dramatically enhanced the accuracy of computer vision-based solutions. However, these solutions only focus on performance and remain insufficient for autonomous driving, as they lack the real-time component. Further, these models have high computational and memory costs, which make them unsuitable for time- and resource-critical applications like autonomous driving. Moreover, solutions like, EfficientNet \cite{tan2019efficientnet} and MLP-mixer \cite{tolstikhin2021mlp} try to improve the overall efficiency of perception solutions by employing simpler and efficient components; however because of dense and high-resolution input, the improvements stay limited.

To obtain rich perception, autonomous vehicles are equipped with multiple sensors, including cameras, LIDAR, and RADAR. Due to higher spatial resolution, camera-based data is considered richer than other sensors, and therefore, even multimodel perception solutions for autonomous driving rely heavily on camera data. However, camera-based perception solutions face multiple challenges, preventing them from being suitable for autonomous driving. 1) Greater resources are required to process higher-resolution cameras, resulting in higher processing time. 2) Cameras mounted on the moving vehicle make the scene dynamic, with traffic objects moving with high relative velocity. This high relative velocity results in motion blur, which impacts the accuracy. 3) The performance of the these solutions drops dramatically in low-light conditions, where a large amount of information is lost due to the absence of light. This can be addressed by reducing the shutter speed of the camera, allowing more light to be captured; however, it intensifies motion blur. 4) Autonomous vehicles should be able to operate in remote as well as public places, and camera images captured in public places raise privacy concerns.

Recently, event cameras surfaced as an efficient alternative to RGB cameras. Unlike their classical counterparts, event cameras capture an event when an intensity change at a particular pixel exceeds a threshold. Event cameras have high dynamic range, negating motion blur and blind time, which exist in RGB cameras between the frames. Further, due to their event-based working principle, event cameras are more effective in low-light conditions. Moreover, since event cameras do not capture RGB images, it is almost impossible to recover identities from event streams, and hence, they can be used in public places without privacy concerns. However, compared to RGB cameras, event cameras are quite recent, and hence, the research field is still in the early development phase. 



Recently proposed, recurrent vision transformers (RVT) \cite{gehrig2023recurrent} try to bridge the gap between the performance of RGB and event camera-based solutions. RVT \cite{gehrig2023recurrent} uses a ViT-based \cite{dosovitskiy2020image} backbone, which includes an LSTM \cite{hochreiter1997long} block at the end of each stage. To efficiently utilize information from different spatial regions, RVT \cite{gehrig2023recurrent} uses multi-axis attention, which results in better performance and throughput compared to previous methods. However, the attention-based designs have a larger memory footprint and higher computational cost. Further, the multi-axis attention mechanism divides the event frame into patches, which causes immediate neighboring pixels to fall in different patches. Although it provides the paths for information flow between patches, these are indirect paths. This unnatural division and indirect paths result in a performance drop and decrease model efficiency. To resolve these issues, we propose a novel event object detection backbone composed of an event-tailored feature extractor module followed by multiple MetaFormer \cite{yu2022metaformer} like blocks. The key novelty of our backbone is Event Progression Extractor (EPE), which is tailored for event data. Unlike prior approaches that mix spatial and temporal features simultaneously, EPE enhances per-pixel event progression features first, preserving fine-grained motion cues before spatial features dominate. The MetaFormer blocks use RepMixer as a building block and convolution-based token and channel mixers. We also employ a RepMixer-based tokenizer to avoid patching and use train-time overparametrization to improve accuracy and efficiency.

To evaluate the efficiency and performance of our model, we conduct a series of experiments and benchmark it on well-established event camera-based object detection datasets. We present both qualitative and quantitative results underscoring the performance, efficiency,  generalizability, and scalability of the proposed model. \textbf{The list of major contributions of this work is as follows:}
\begin{enumerate}
\item We propose a novel and efficient backbone for event-based object detection using MetaFormer \cite{yu2022metaformer} blocks with convolution-based components, LSTMs \cite{hochreiter1997long} and an event-tailored feature extractor.

\item We propose a novel Event Progression Extractor as an event-tailored feature extractor  to enable temporal feature enrichment at the early stage, which is necessary to fully exploit event progressions.

\item We perform extended experiments to evaluate our proposed model and compare it against state-of-the-art on well-established benchmarks.

\item Our proposed model becomes the fastest DNN-based architecture to date, with a $14\%$ reduction in inference time compared to the current state-of-the-art on benchmark datasets.

\item We conduct a comprehensive ablation study on choice of tokenizer, channel mixer and token mixer.
\end{enumerate}

\section{Related Work}
\label{sec:related}

Event-based object detection literature and research can be broadly categorized into two types; (1) Event Data Representations and (2) Deep Neural Networks (DNNs). In this section, we will discuss these categories in detail.

\subsection{Event Data Representations}
\label{subsec:related_event_data_rep}

The sparsity of events data poses challenges when interfacing with neural networks designed for frame-based data. These neural networks inherently demand dense representations or 2D images as input. To address this issue, \cite{cook2011interacting, munda2018real, rebecq2019events} devise methods for the conversion of asynchronous event data into compact representations suitable for subsequent neural network computation. A widespread approach is to turn event streams into gray or color images and then use vision-based deep neural networks to process them.
Rebecq et al. \cite{rebecq2019high} introduce a UNet-based \cite{ronneberger2015u} recurrent architecture for the direct reconstruction of gray images from events data. However, such approaches cause a computational overhead due to the conversion of events to images.

Alternatively, to harness sparse events directly, several handcrafted representations are proposed. A simplistic approach involves the accumulation of events at each spatial location (pixel) over time, resulting in histograms \cite{cook2011interacting, maqueda2018event, rebecq2016evo}. However, this naive approach neglects the temporal properties inherent in events data. To solve this, innovative solutions like 2D time surfaces are proposed to exploit temporal resolution by capturing the timestamp of the most recent event at each pixel \cite{lagorce2016hots}. Building upon this concept, \cite{sironi2018hats} proposes Histogram of Averaged Time Surfaces (HATS), a robust event-based representation using local memory units.

To address both spatial and temporal information, widespread approaches focus on the creation of 3D voxel grids or event volumes, where each cube corresponds to a specific pixel and time interval. Perot et al. \cite{perot2020learning} propose the generation of event cubes with micro time bins and polarity information, subsequently utilizing ConvLSTM \cite{shi2015convolutional} for object detection. Similarly, Gehrig et al. \cite{gehrig2019end} present an end-to-end learning methodology, transforming event streams into grid-based representations termed Event Spike Tensor (EST) through a sequence of differential operations. This approach demonstrates superior performance in optical flow \cite{ zhu2018ev} and object recognition tasks \cite{lagorce2016hots, orchard2015hfirst}. Another noteworthy end-to-end method is MatrixLSTM \cite{cannici2020differentiable}, employing a grid of LSTM cells with shared parameters, achieving state-of-the-art results on N-Cars \cite{sironi2018hats}, N-Caltech \cite{orchard2015converting} image classification datasets, and the MVSEC optical flow estimation benchmark \cite{zhu2018multivehicle}. ERGO \cite{zubic2023chaos} proposes a 12 channel event representation; ERGO-12 by optimizing over a combination of representations using Gromov-Wasserstein Discrepancy (GWD) metric.




\subsection{DNNs for Event-based Object Detection}
\label{subsec:deep_neural_net}

Deep Neural Networks (DNNs) are favored for object detection, due to their remarkable accuracy in various vision tasks. \cite{rebecq2019high} proposes a recurrent UNet \cite{ronneberger2015u} architecture for reconstructing high-quality images/videos from event streams, proving effective in downstream computer vision tasks, particularly classification and visual-inertial odometry. Li et al., \cite{li2019event} introduce a joint detection framework, combining spatial and temporal features through a CNN architecture and synchronizing modalities with a CNN-SNN model. It fine-tunes YOLOv3 \cite{redmon2016you} on the DDD17 vehicle detection dataset, and performs well in challenging illumination conditions as well.

RED \cite{perot2020learning} introduces a ConvLSTM architecture for extracting rich spatial and temporal features along with a SSD head \cite{liu2016ssd} for detection. ASTMNet \cite{li2022asynchronous}, an end-to-end asynchronous spatio-temporal memory network, outperforms existing methods on Gen1 \cite{de2020large}and 1MPx \cite{perot2020learning} datasets, but faces challenges with memory complexity. Embracing the sparsity of event cameras, recent transformer-based models, such as Event Transformer (EvT) \cite{sabater2022event} and a vision transformer (ViT) \cite{wang2022exploiting}, demonstrate efficiency in classification tasks. RVT \cite{gehrig2023recurrent}, a hierarchical recurrent vision transformer, integrates multi-axis attention and LSTMs \cite{hochreiter1997long} for event-based object detection, showcasing performance on automotive datasets. LEOD \cite{wu2024leod} tackles event object detection as a weakly and semi-supervised problem with self-training, avoiding the need for dense training data annotation. \cite{zubic2024state} builds on recurrent vision transformers and replaces LSTM layers with state-space models to achieve faster training. GET-T \cite{peng2023get} proposes group token event representation and uses a transformer-based architecture to achieve superior performance.  

DNNs have been pivotal in advancing event-based object detection, with recent transformer-based approaches and hierarchical recurrent vision transformers presenting promising strides in accuracy on diverse datasets. However, there is still room for improvement, specially in terms of efficiency as recent approaches use ViT \cite{dosovitskiy2020image} based architecture which does not efficiently use spatial and temporal priors.
\begin{figure*}[!t]
    \centering
    \includegraphics[width=\textwidth]{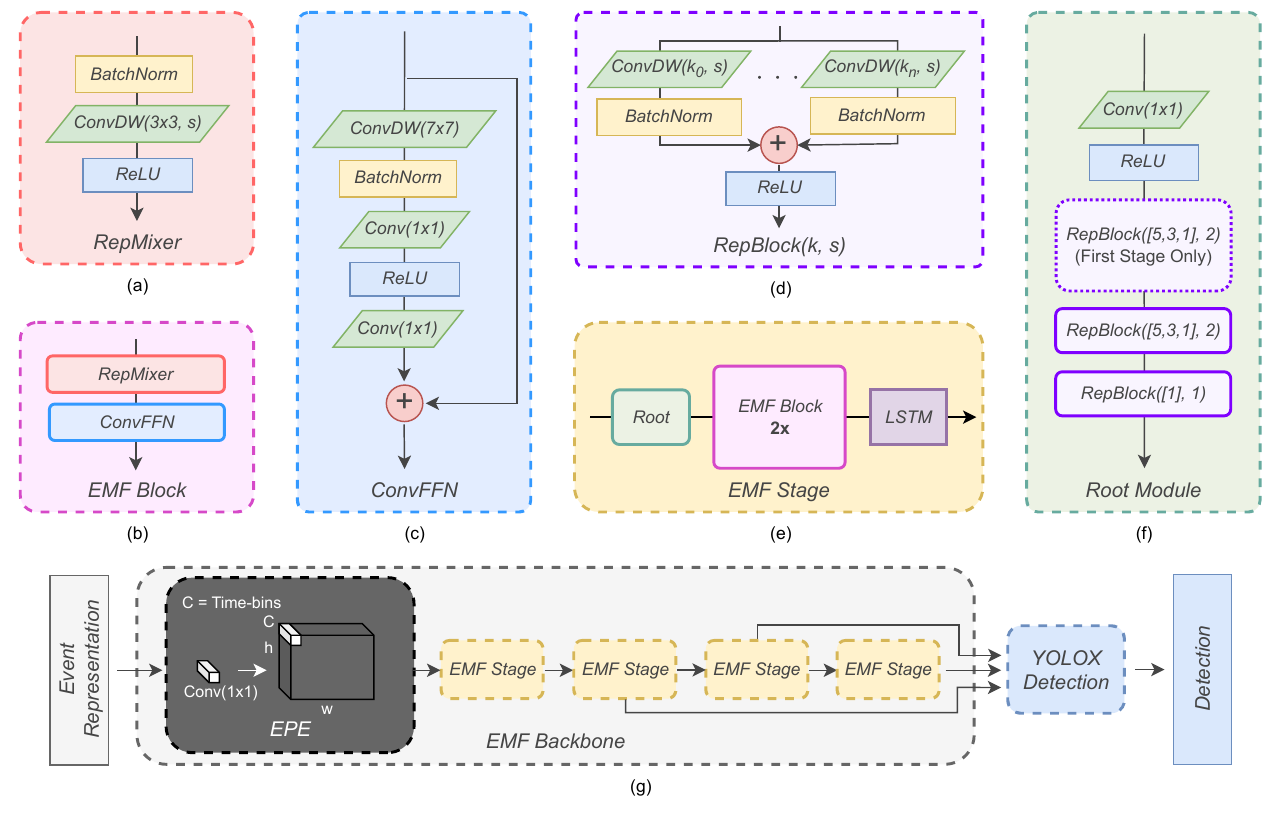}
    \caption{Shows the detailed architecture of the proposed EMF backbone, its components (a-f) and usage in event-based object detection pipeline (g).}
    \label{fig:emf}
\end{figure*}

\section{Method}

This section presents the end-to-end pipeline of our \textit{Event Meta Former} (EMF) architecture. In the first step, it converts events into a rich event representation (Sec. \ref{subsec:method_event_data_rep}). Our novel \textit{Event Meta Former} backbone processes these representations to generate high-level classification and location features (Sec. \ref{subsec:method_backbone}), which are finally sent to a YOLOX detection framework to predict the object bounding boxes and their respective classes (Sec. \ref{subsec:method_detection}).  

\subsection{Event Data Representation}\label{subsec:method_event_data_rep}
The output of an event camera is a sequence of events, with each of the form,
\begin{equation}
    \begin{split}
        e_i &= (x_i, y_i, p_i, t_i), \\
        x &\in [0, W], y \in [0, H], \\
        p &\in \{-1, 1\},
    \end{split}
\end{equation}

\par where $W$ and $H$ are the width and height of the event frame respectively, $p$ is polarity and $t$ is the timestamp of the asynchronous event. Modern deep learning architectures typically require input to be in the form of discrete 2D/3D volume.  To utilize these frame-based architectures, it is important to convert the steam of events into 3D input volume which can be easily processed by common deep neural network components, like convolutions.
    
Following our baseline \cite{gehrig2023recurrent}, we use \textit{Stacked Histograms} as event representation, with $nbins = 10$ and $dt = 50ms$. \textit{Stacked Histograms} extend \textit{Histogram of Events}, which creates histograms of positive and negative events per pixel, by using multiple time-bins to preserve motion information within the time-frame. \textit{Stacked Histograms} divide the event stream into spatially and temporarily discretized event volumes of predefined time duration and spatial resolution. Each volume is a 4D tensor, i.e., $(P, T, H, W)$, where $P = 2$ to cater negative and positive events separately, and $T$ preserves the motion information within the time-frame by dividing it further into time bins, with a standard of $10$. To prepare the volume as input to the network, it is reshaped in a 3D tensor by merging the first two dimensions, i.e., $(PT, H, W)$.

\subsection{Event Meta Former Backbone}\label{subsec:method_backbone}

Correlated events forming object contours in event data are major clues for an object detector to precisely locate them. Large convolution kernels can help capture this correlation and, hence, achieve better accuracy. Moreover, estimating the progression of events over time combined with large convolution kernels can provide vital information to detect objects. Furthermore, compared to attention, convolutions allow better information flow due to inherent positional priors. To this extent, we propose a novel recurrent, convolution-based backbone for event-based object detection, named \textit{Event Meta Former}. Our proposed backbone includes \textit{Event Progression Extractor} and four \textit{EMF Stages} in total, with each having a \textit{Root Module} as a tokenizer at the start, followed by two \textit{EMF Blocks} and an LSTM layer. Before passing samples to \textit{EMF Stages}, our \textit{EMF} backbone employs \textit{Event Progression Extractor} (EPE) layer in the form of point-wise convolution on event-bins to capture progression of event over time. Our \textit{EMF Block} is a Metaformer \cite{yu2022metaformer} like block which uses \textit{RepMixer} \cite{vasu2023fastvit} as token mixer and \textit{ConvFFN} \cite{vasu2023fastvit} as channel mixer.

The EPE module takes event representations and enriches temporal features by extracting event progressions before being overwhelmed by spatial features. The \textit{Root Module} downsamples the feature-volume, using \textit{RepMixer} with overlapping kernels. The \textit{EMF Block} enriches both spatial and temporal features within a sample; the LSTM modules at the end of each stage enable the network to capture temporal information between sequence samples. Fig. \ref{fig:emf} shows the detailed architecture of our proposed backbone, along with its usage in the event-based object detection pipeline.

\subsubsection{Event Progression Extractor}

In a CNN-based backbone for images, large kernels are used at the start of the backbone to exploit the relation between neighboring pixels. However, this strategy does not work well with event representations, as the network gets overwhelmed by the higher magnitude of gradients from the spatial dimension and not able to learn the temporal feature properly. To resolve this issue, we propose to use our EPE module at the start of the backbone. The EPE module contains point-wise convolutions which only focus on the temporal relations and enriches them, resulting in high-level temporal features.

\subsubsection{RepBlock}


The RepVGG \cite{ding2021repvgg} uses repetitive convolutions of multiple kernel sizes on the same input and merges the output by adding them. In this way, it can capture features at different scales and achieve higher throughput by merging multiple convolutions into one for inference. FastViT \cite{vasu2023fastvit} extends the idea and uses Depth-Wise separable convolutions instead of normal convolutions to allow spatial connection only, this setup enables local feature enrichment with higher throughput. Fig. \ref{fig:emf}d shows the components of the \textit{RepBlock} which follows the same idea.

\subsubsection{Root Module}

The goal of the \textit{Root Module} is to downsample the spatial dimensions, expand channels, and gather information across both dimensions to prepare it for further processing. We use a point-wise convolution at the start to expand the channel dimension, followed by multiple \textit{RepBlocks}. The earlier \textit{RepBlocks}, use stride of $2$ with large kernels to enrich spatial connections and downscale the feature-maps. While the later \textit{RepBlock} uses $1\times1$ Conv with the stride of $1$ to focus on channel-wise connections. Conventionally, the first stage of backbone downscales the feature-maps by the factor of $4$ while the other stages downscales them further, each by the factor of $2$. To achieve this, we use \textit{RepBlock} with stride of $2$, twice in first stage and once in the rest. Fig. \ref{fig:emf}f shows the architecture of the \textit{Root Module}.

\subsubsection{Token Mixer}

The goal of a token mixer in a Metaformer-based network is to learn local features. \cite{dosovitskiy2020image} uses attention as a token mixer, however; more efficient and light alternatives are available which produce similar performance \cite{yu2022metaformer} i.e., 2D average pooling, and MLPs. We use \textit{RepMixer} as a token mixer similar to \cite{vasu2023fastvit} as it is simple, performant and efficient thanks to Depth-Wise separable convolution.

\subsubsection{Channel Mixer}

The goal of a channel mixer is to capture information based on features of a particular location. We use ConvFFN \cite{vasu2023fastvit} as a channel mixer; it uses a large depth-wise separable convolution to gather neighboring information, followed by back to back point-wise convolutions to extract features from different channels. Since, all convolutions used in the module are 2D, it achieves higher throughput.

\subsection{The Detection Framework}
\label{subsec:method_detection}

YOLOX \cite{ge2021yolox} is a well established detection framework employed widely by recent event-based object detection techniques. In contrast to its predecessors \cite{redmon2016you, redmon2018yolov3} YOLOX \cite{ge2021yolox} uses anchor-free design, which performs object detection in an end-to-end fashion by predicting, label and bounding box per-pixel instead of per anchor. This approach simplifies the architecture, improves the efficiency, and decreases the training and inference time. YOLOX \cite{ge2021yolox} performs classification and regression using two separate branches, allowing the network to learn attribute specific features from common feature-maps. This, along with \textit{simOTA} label assignment strategy, results in a significant performance boost. The loss function of the YOLOX detection head \cite{ge2021yolox} is given by,

\begin{equation}
    L = L_{cls} + \lambda L_{reg},
\end{equation}

\par where $\lambda$ is the balancing factor.
\begin{table}[t]
  \caption[Datasets table]{Summary of Event-Based Object Detection Datasets.}\label{tab:datasets}
  \centering
  \renewcommand{\arraystretch}{1.2}
    \resizebox{\linewidth}{!}{
  \begin{tabular}{l|cccc|c}
    \hline
      \textbf{Dataset} & \textbf{Year} & \textbf{Resolution} & \textbf{Classes} & \textbf{Size} & \textbf{Labels} \\
    \hline
      
      \multirow{2}{*}{
        Gen1 \cite{de2020large} 
      }
      & \multirow{2}{*}{2020} & \multirow{2}{*}{304x240} & \multirow{2}{*}{2} & \multirow{2}{*}{39.0 hrs} & \multirow{2}{*}{Cars, Pedestrian} \\
       & & & & &
      \\
      \hline
      \multirow{4}{*}{
        1 Mpx \cite{perot2020learning} 
      }
      & \multirow{4}{*}{2020} & \multirow{4}{*}{1280x720} & \multirow{4}{*}{6} & \multirow{4}{*}{14.6 hrs} & \multirow{4}{*}{
      \makecell{
      Car, Pedestrian,\\
      Two-wheelers, \\
      Truck, Van, \\
      Traffic-light}
      }  \\
       & & & & & 
      \\
      & & & & &
      \\
      & & & & &
      \\
      \hline
  \end{tabular}
  }
\end{table}

\section{Experimental Setup}
This section contains the details of our experimental setup. We start with listing details of datasets used in this work, followed by evaluation metric used to test our proposed architecture and hardware setup used to perform training, testing, and inference time calculations.

\begin{table*}[t]
    \caption[Results table]{Object Detection Benchmarks results on the Gen1 \cite{de2020large} and 1Mpx \cite{perot2020learning} event-based automotive detection datasets. The baseline results are reported from \cite{gehrig2023recurrent}. Evaluations are done on a single RTX3090 GPU with 1 worker, 128GB RAM and 1 sample per batch. The \textbf{best} results are in bold, while the \underline{second best} results are underlined.}
    \label{tab:results_test}
    \centering
    \renewcommand{\arraystretch}{1.2}
    \resizebox{\linewidth}{!}{
        \begin{tabular}{l|l|l|cc|cc|c|c}
            \hline
            \multicolumn{1}{c|}{
                \multirow{2}{*}{
                    \textbf{Method}   
                }
            }
            &&
            & \multicolumn{2}{c|}{\textbf{Gen1}}
            & \multicolumn{2}{c|}{\textbf{1 Mpx}} & Avg.
            & \\
            \cline{4-5} \cline{6-7} 
            & Backbone & Detection Head & mAP & Inf. (ms) & mAP & Inf. (ms) & mAP & Params (M)\\
            \hline
            ASTMNet \cite{li2022asynchronous} & (T)CNN + RNN & SSD & 46.7 & 35.6 & \underline{48.3} & 72.3 & 47.5 & \textgreater 100  \\
            S5-ViT-B \cite{zubic2024state} & Transformer + SSM & YOLOX & 47.4 & 32.0 & 47.2 & 45.5 & 47.3 & 18.2
            \\
            RED \cite{perot2020learning} & CNN + RNN & SSD & 40.0 & 16.7 & 43.0 & 39.3 & 41.5 & 24.1 \\
            GET \cite{peng2023get} & Transformer + RNN & YOLOX & 47.9 & 16.8 & \textbf{48.4} & 18.2 & \textbf{48.2} & 21.9 \\
            RVT-B \cite{gehrig2023recurrent} & Transformer + RNN & YOLOX & 47.2 & 11.2 & 47.4 & 11.8 & 47.3 & 18.5
            \\
            \hline
            RVT-S \cite{gehrig2023recurrent} & Transformer + RNN & YOLOX & 46.5 & 10.4 & 44.1 & 10.9 & 45.3& 9.9 \\
            LEOD-RVT-S \cite{wu2024leod} & Transformer + RNN & YOLOX & \underline{48.7} & 10.4 & 46.7 & 10.9 & \underline{47.7} & 9.9 \\
            RVT-T \cite{gehrig2023recurrent} & Transformer + RNN & YOLOX & 44.1 & \underline{10.3} & 41.5 & \underline{10.5} & 42.8 & 4.4 \\
            \rowcolor{lgray}
            EMF (ours) & Metaformer + RNN & YOLOX & \textbf{49.1} & \textbf{9.1} & 46.3 & \textbf{9.3} & \underline{47.7} & 14.9 \\
            \hline
        \end{tabular}
    }
\end{table*}

\subsection{Datasets}
\label{sec:datasets}
Datasets are a key aspect of deep learning, as the quality and abundance of data has a major impact on the performance of these models. In this work, we used two widely accepted event-based object detection datasets, i.e., Prophesee Gen 1 and Prophesee 1 Mpx dataset. Tab. \ref{tab:datasets} shows summary of these datasets.

\subsubsection{Prophesee Gen1 Automotive Detection Dataset}

Prophesee Gen1 is one of the largest event-based automotive dataset \cite{de2020large} released in $2020$. In comprises more than $39$ hours of recordings captured with the $304\times240$ Gen1 ATIS sensor \cite{simon2016event}. These recordings include open road and various driving scenarios ranging from urban, highway, suburbs and countryside scenes, captured in changing lighting and weather conditions. 

The annotation is done manually using gray level estimation feature of the ATIS camera. Two classes, cars and pedestrians, are labeled considering their importance in autonomous driving scenarios. In total, the dataset contains around $256K$ bounding box annotations with approximately $228K$ cars and $28K$ pedestrians \cite{de2020large}.
We follow the evaluation protocol of Gen1 dataset \cite{perot2020learning} in our experiments. All the bounding boxes with a side length of less than $10$ pixels and a diagonal of less than $30$ pixels are removed.

\subsubsection{Prophesee 1 Megapixel Automotive Detection Dataset}

Prophesee 1 Megapixel \cite{perot2020learning} is the first real-world high resolution event-based automotive dataset to date. The dataset is recorded using a 1 Mpx events camera \cite{finateu20205} with a combined recorded data of $14.65$ hours. These recordings are split into $11.19$ hours for training, $2.21$ hours for validation and $2.25$ hours for testing. Recordings are captured during the daytime in various scenarios, and under changing lighting and weather conditions. In all the recordings, both the event and frame camera are mounted behind the windshield of the car. A total of $25M$ bounding boxes are annotated, belonging to seven classes. Labels are first extracted from an RGB camera and then transferred to the event camera coordinates by using homography.

In our experiments on 1Mpx dataset \cite{perot2020learning}, we follow the evaluation protocols given with the dataset. The input event representation resolution is downsampled by a factor of $2$ ($640\times360$) and all the bounding boxes with a side length of less than $20$ pixels and a diagonal of less than $60$ pixels are also removed. To be consistent with previous research, only three classes, i.e., cars, pedestrians and two-wheelers are used out of seven classes in the dataset.

\subsection{Evaluation Criteria}

\textit{Mean Average Precision} is a standard evaluation metric for object detection in both RGB camera and event camera domain. We use the COCO evaluation API \cite{lin2014microsoft} along with protocols proposed by RED \cite{perot2020learning}. In the results, we report $mAP$ short for $mAP[50-95]$, which indicates mean average precision values at different IOU thresholds, ranging from $50\%$ to $95\%$.

\subsection{Training and Evaluation Settings}

For our experiments, we use the similar training settings as RVT \cite{gehrig2023recurrent}. We do mix precision training, spanning a minimum of 400K steps. For optimization, we utilize the ADAM optimizer \cite{kingma2014adam} in combination with 1 cycle learning rate schedule \cite{smith2019super}. 
Also, we employ a mixed batching strategy, which applies backpropagation through time (BPTT) to half of the samples and truncated BPTT (TBPTT) for the rest.

The training on the Gen1 dataset \cite{sironi2018hats} is carried out using a batch size of $8$, a sequence length of $21$, and a learning rate of $2\times10^{-4}$ on a single A100 GPU. 
For the 1 Mpx dataset, we employ a larger batch size of $24$, a shorter sequence length of $5$, and a slightly higher learning rate of $3.5\times10^{-4}$. 

The evaluation results are reported on test set for both the Gen1 dataset and the 1 Mpx dataset \cite{perot2020learning}. The evaluation is done on a single RTX3090 GPU, with a number of workers and batch size of $1$. For ablation study, we use validation sets of Gen1 dataset and evaluation batch size of $8$ to expedite the experiments.

\subsection{Inference Time Calculation}
To calculate inference time, we evaluate the models on RTX 3090 GPU with a batch size of $1$. The inference time is calculated as the mean difference of time when the image tensor, already loaded in the GPU memory, is passed to the model for inference and the time when the detection head returns the output. For fair comparison, all inference times reported in this work are of non JIT-compiled models.

\begin{figure*}[!t]
    \centering

    \begin{tabular}{c c}
        \textit{Gen 1} \cite{de2020large} \hspace{0.18\textwidth} & \hspace{0.18\textwidth} \textit{1 Mpx} \cite{perot2020learning}
    \end{tabular}

    \vspace{0.5em}

    \subfloat{\raisebox{2.5em}{\rotatebox[origin=t]{90}{{\small \textit{EMF}}}}\;\;}
    \subfloat{\includegraphics[width=0.156\textwidth]{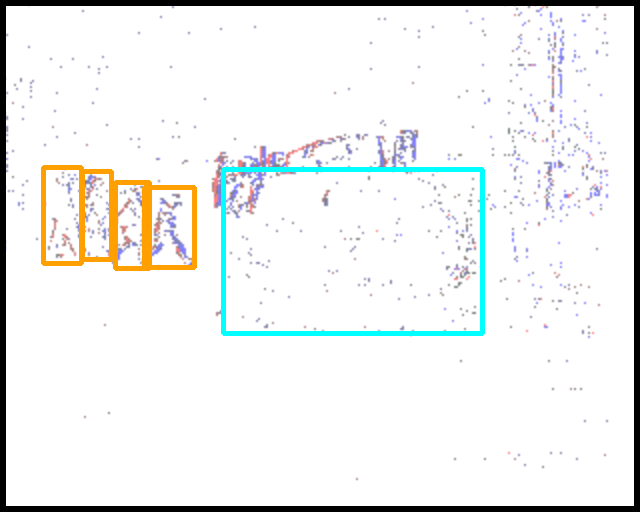}\hspace{0.004\textwidth}}
    \subfloat{\includegraphics[width=0.156\textwidth]{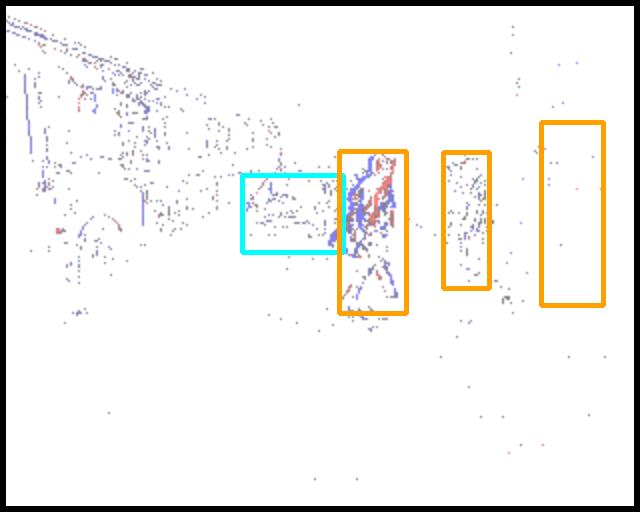}\hspace{0.001\textwidth}}
    \subfloat{\includegraphics[width=0.156\textwidth]{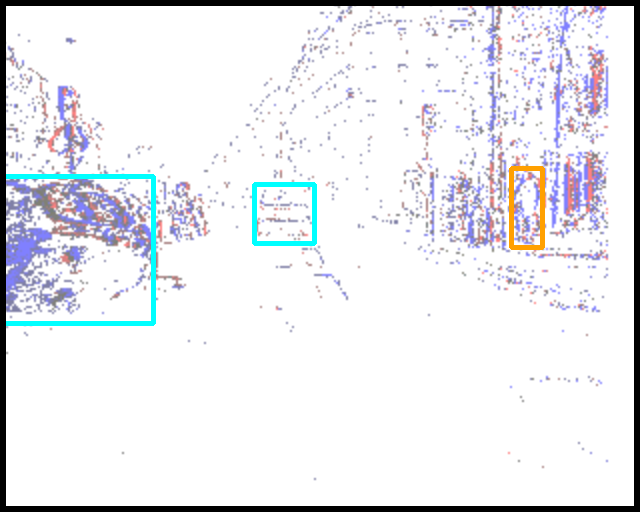}\hspace{0.005\textwidth}}
    \subfloat{\includegraphics[width=0.156\textwidth]{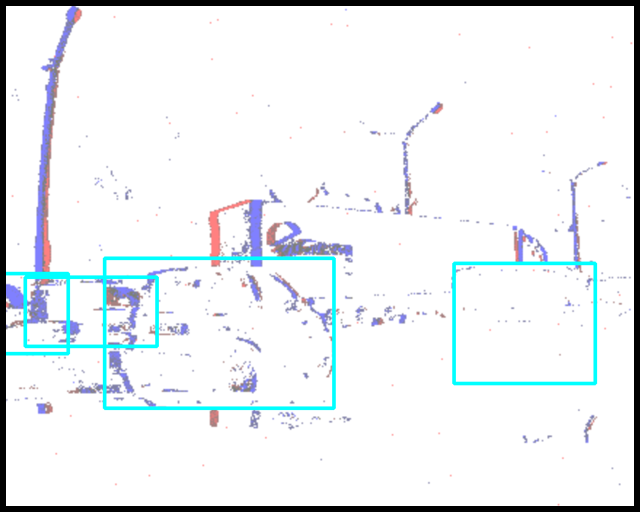}}
    \subfloat{\includegraphics[width=0.156\textwidth]{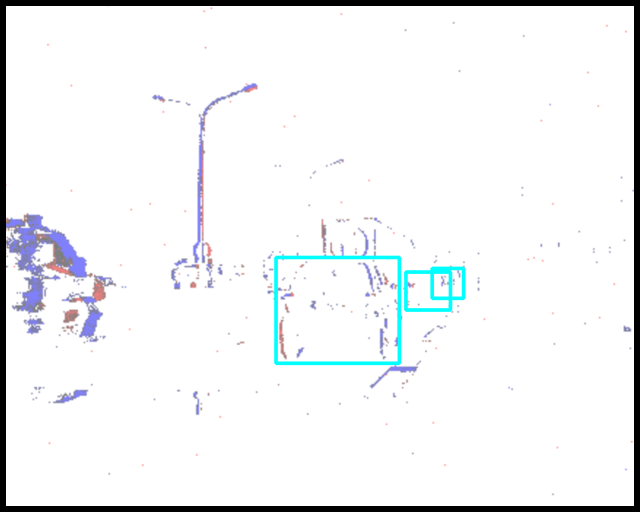}\hspace{0.005\textwidth}}
    \subfloat{\includegraphics[width=0.156\textwidth]{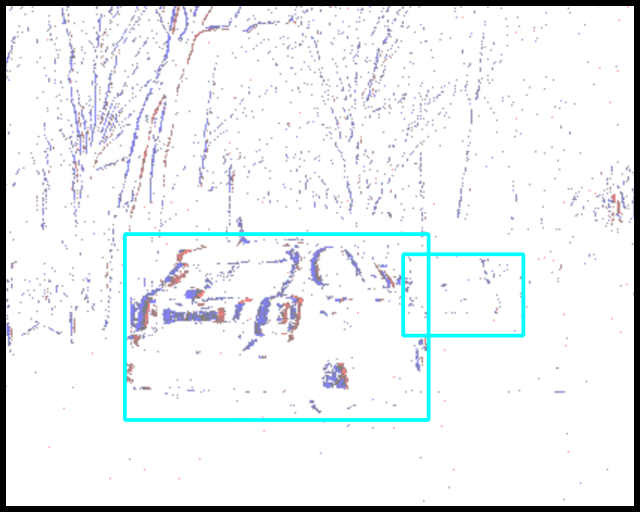}}

    \vspace{0.015em}

    \subfloat{\raisebox{2.5em}{\rotatebox[origin=t]{90}{{\small \textit{GT}}}}\;\;}
    \subfloat{\includegraphics[width=0.156\textwidth]{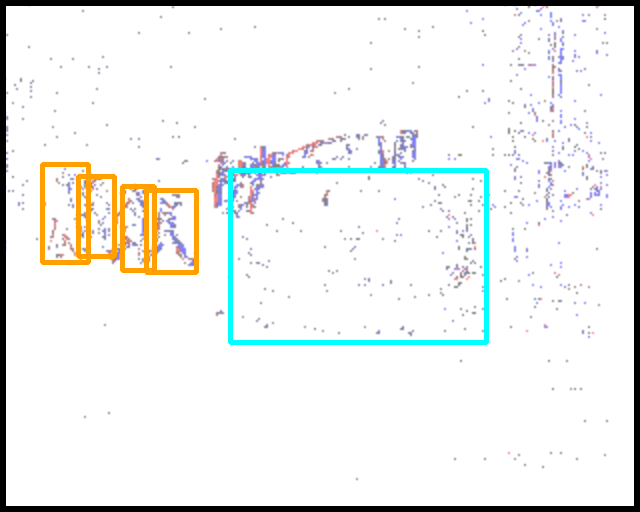}\hspace{0.004\textwidth}}
    \subfloat{\includegraphics[width=0.156\textwidth]{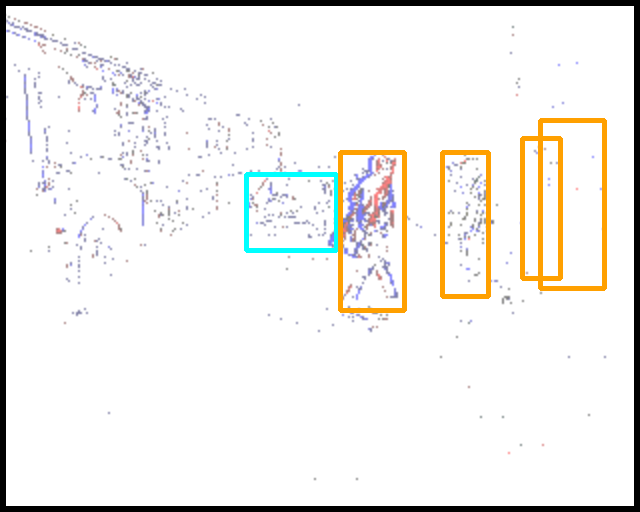}\hspace{0.001\textwidth}}
    \subfloat{\includegraphics[width=0.156\textwidth]{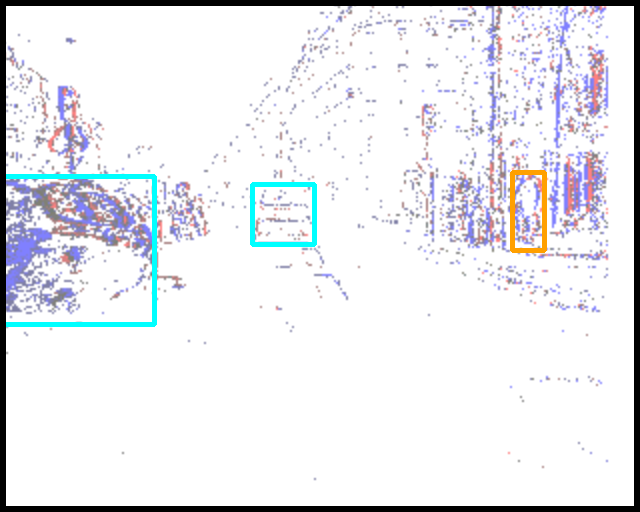}\hspace{0.005\textwidth}}
    \subfloat{\includegraphics[width=0.156\textwidth]{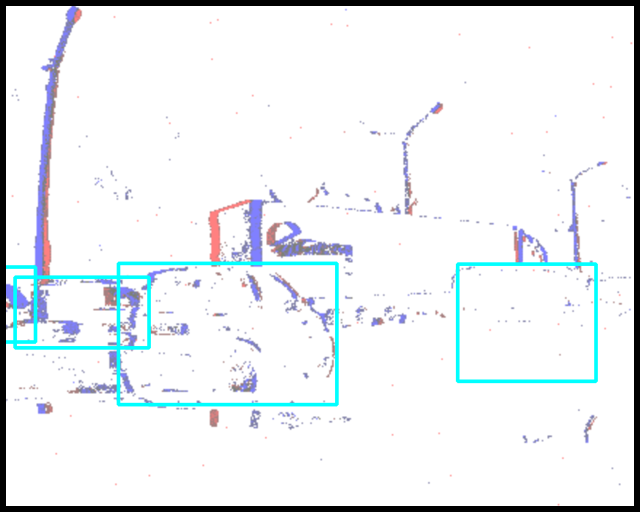}}
    \subfloat{\includegraphics[width=0.156\textwidth]{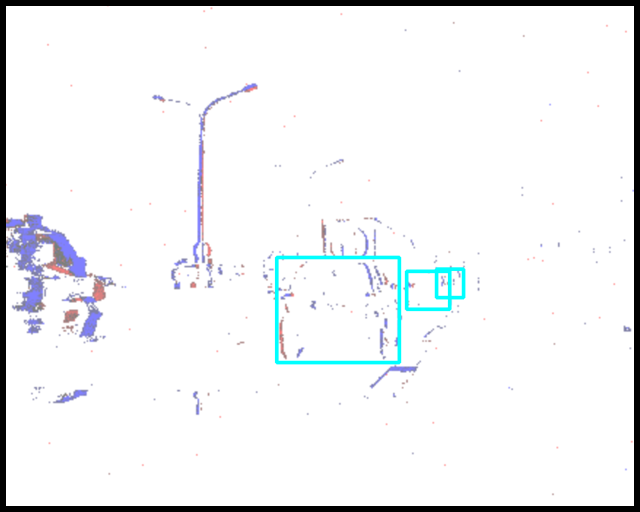}\hspace{0.005\textwidth}}
    \subfloat{\includegraphics[width=0.156\textwidth]{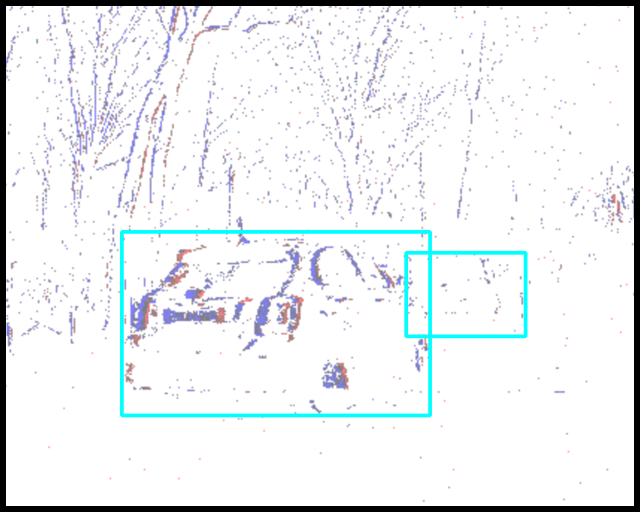}}

    \vspace{0.02em}
        
    \subfloat{\raisebox{2.5em}{\rotatebox[origin=t]{90}{{\small \textit{RVT-B}}}}\;\;}
    \subfloat{\includegraphics[width=0.156\textwidth]{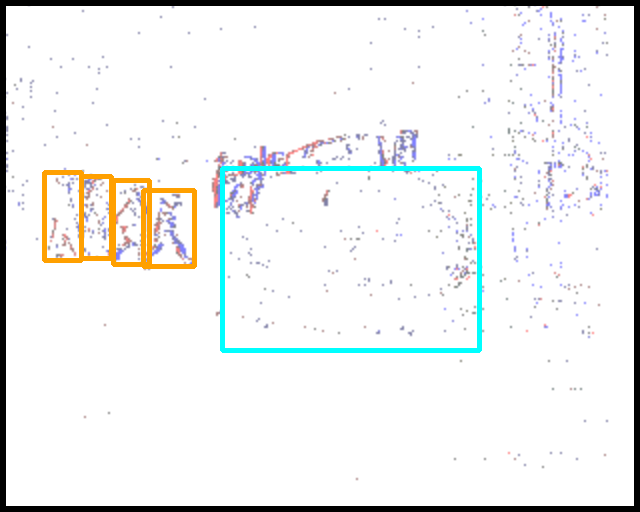}\hspace{0.004\textwidth}}
    \subfloat{\includegraphics[width=0.156\textwidth]{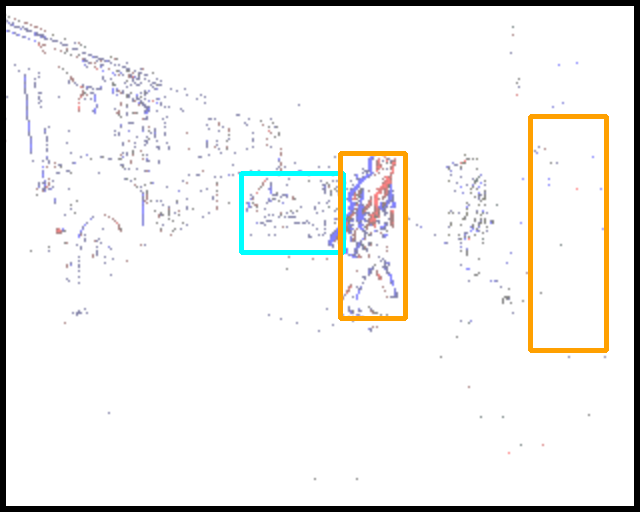}\hspace{0.001\textwidth}}
    \subfloat{\includegraphics[width=0.156\textwidth]{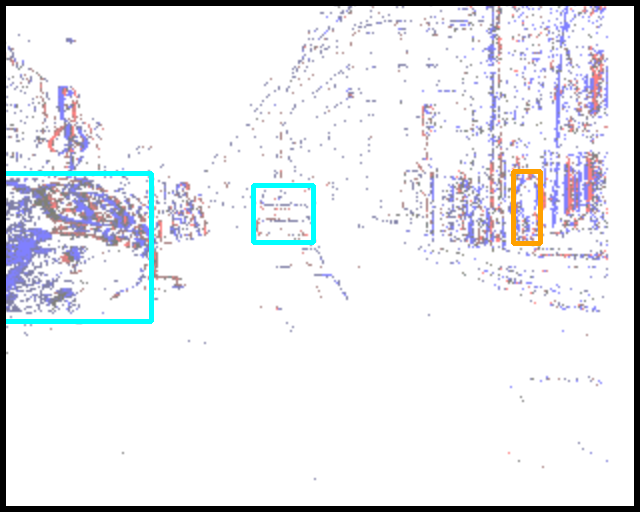}\hspace{0.005\textwidth}}
    \subfloat{\includegraphics[width=0.156\textwidth]{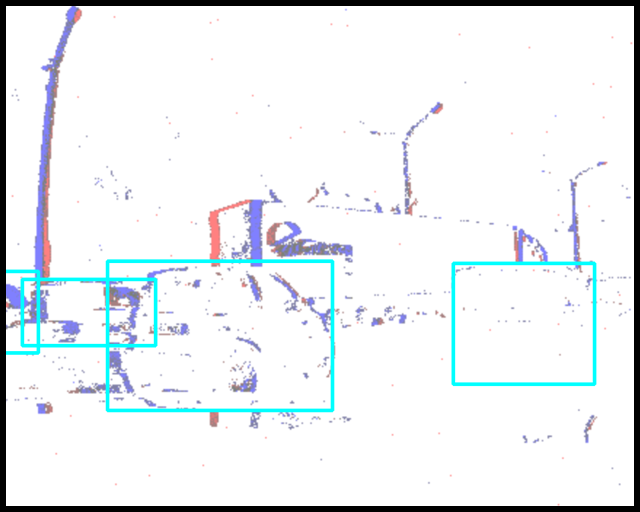}}
    \subfloat{\includegraphics[width=0.156\textwidth]{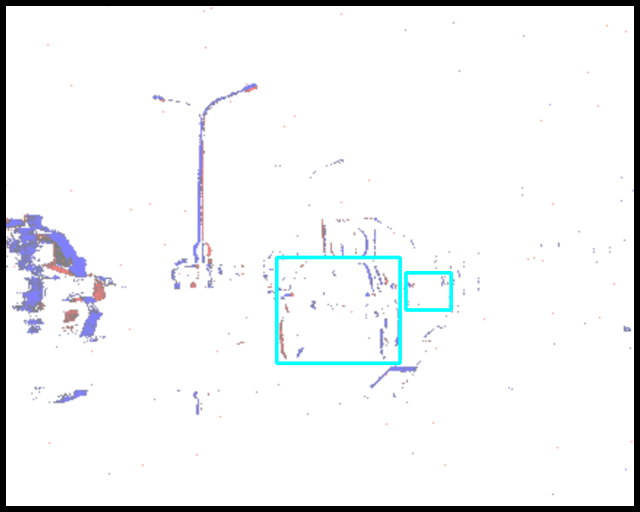}\hspace{0.005\textwidth}}
    \subfloat{\includegraphics[width=0.156\textwidth]{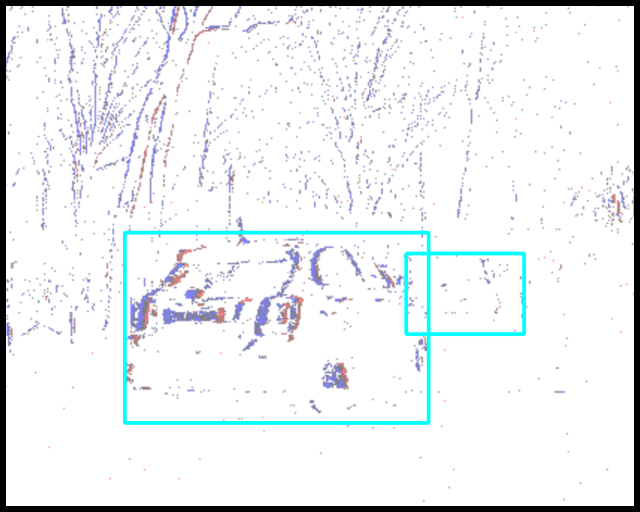}}
    
    \caption{Qualitative comparison of EMF and RVT-B \cite{gehrig2023recurrent} models against ground-truth (GT) on 1Mpx dataset. The cyan bounding boxes represent cars, while the orange bounding boxes represent pedestrians. The first three columns are from the Gen1 dataset, while the last three columns contain samples from the 1Mpx dataset.}
    \label{fig:qual_res}
\end{figure*}
\section{Results}
To evaluate our proposed approach and compare it against the state-of-the-art, we conduct multiple experiments. We compare our proposed \textit{EMF} with state-of-the-art on well established benchmarks, followed by cross dataset evaluations and progressive fine-tuning experiments to test the ability of our proposed model to generalize and scale with the abundance of data. We also perform qualitative comparison with the state-of-the-art and present ablation study at the end of the section.

\subsection{Comparison with the State-of-the-art}

RVT \cite{gehrig2023recurrent} uses an \textit{RNN $+$ Transformer} backbone architecture to achieve state-of-the-art object detection accuracy on event-based automotive datasets. We compare \textit{EMF} with state-of-the-art detectors to test its relative performance and efficiency. Tab. \ref{tab:results_test} shows the results of these experiments. \textit{EMF} outperforms RVT-B, on average, by $0.4\text{ } mAP$ with $1.9\text{ } mAP$ improvement over Gen1 automotive dataset. Further, \textit{EMF} achieves this performance with $20\%$ lesser inference time and $19\%$ lesser model parameters. Compared to RVT-S, which is a smaller and faster version of RVT, \textit{EMF} achieves $2.4$ better $mAP$ with $14\%$ lesser inference time. Compared to LEOD \cite{wu2024leod}, which uses improved datasets with less noisy samples, \textit{EMF} achieves similar performance with $14\%$ lesser inference time, showing its robustness to noisy samples. GET \cite{peng2023get} uses a richer event representation, which contributes to its stronger performance ($0.5\text{ }mAP$ higher compared to EMF). Despite this, EMF achieves: $47\%$ lower inference time than GET and $1.2\text{ }mAP$ higher on Gen1. Besides these performances, our proposed \textit{EMF} is the fastest DNN-based model to date in event-based object detection, outperforming the fastest version of RVT \cite{gehrig2023recurrent} i.e., RVT-T, with $12\%$ lesser inference time while maintaining a significant performance margin.

\subsection{Cross Dataset Evaluation}
Driving scenarios comprise diverse situations including varying weather, lighting and demography. Traffic object detection methods need to show robustness to these changes in addition to good performance on benchmarks. To test how well our proposed model generalizes to diverse situations and adapts to unseen data, we perform cross dataset evaluations. Tab. \ref{tab:cross} shows the results of these evaluations, where we compare RVT \cite{gehrig2023recurrent} with our proposed \textit{EMF}. It is evident that \textit{EMF} demonstrates superior generalization to unseen data compared to RVT \cite{gehrig2023recurrent}, achieving an average $mAP$ improvement of $1.0$.

\begin{table}[!t]
    \renewcommand{\arraystretch}{1.2}
    \caption{Results of the cross-dataset evaluation experiments.}
    \label{tab:cross}
    \centering
    \resizebox{\linewidth}{!}{
        \begin{tabular}{l|c|c}
             \hline
             \multicolumn{1}{c|}{
                \multirow{2}{*}{
                    \textbf{Method}   
                }
            } 
             & \multicolumn{2}{c}{\textbf{mAP}}
             \\
             \cline{2-3}
              & Train: 1Mpx, Test: Gen1 & Train: Gen1, Test: 1Mpx
             \\
             \hline
            RVT-B \cite{gehrig2023recurrent} & 28.4 & 17.3
            \\
            \rowcolor{lgray}
            EMF (ours) & \textbf{29.9} & \textbf{17.8}
            \\
            \hline
        \end{tabular}
    }
\end{table}

\begin{table}[!t]
    \renewcommand{\arraystretch}{1.2}
    \caption{Results of the progressive fine-tuning. ERGO \cite{zubic2023chaos} uses pre-trained Swin Transformer V2.}
    \label{tab:fine_tune}
    \centering
    \resizebox{0.98\linewidth}{!}{
        \begin{tabular}{l|l|c}
             \hline
             \textbf{Method} & \textbf{Training Strategy} & \textbf{mAP}
             \\
             \hline
            ERGO \cite{zubic2023chaos} & Swin Transformer V2 \cite{liu2022swin} $\rightarrow$ Gen1 & 50.4
            \\
            RVT \cite{gehrig2023recurrent} & 1Mpx $\rightarrow$ Gen1 & \textbf{50.8}
            \\
            \rowcolor{lgray}
            EMF (ours) & 1Mpx $\rightarrow$ Gen1 & \textbf{50.8}
            \\
            \hline
            ERGO \cite{zubic2023chaos} & Swin Transformer V2 \cite{liu2022swin} $\rightarrow$ 1Mpx & 40.6
            \\
            RVT \cite{gehrig2023recurrent} & DSec $\rightarrow$ 1Mpx & 32.4
            \\
            \rowcolor{lgray}
            EMF (ours) & DSec $\rightarrow$ 1Mpx & \textbf{42.5}
            \\
             \hline
        \end{tabular}
    }
\end{table}

\begin{table*}[!t]
    \renewcommand{\arraystretch}{1.2}
    \caption{Ablation study of different Metaformer\cite{yu2022metaformer} and ViT\cite{dosovitskiy2020image} architectures on Gen1 dataset. The inference time is calculated with 8 samples per batch on a single RTX 3090.}
    \label{tab:ablation}
    \centering
    \resizebox{\linewidth}{!}{
        \begin{tabular}{l|cclll|ccc}
             \hline
             \textbf{Model} & \textbf{Patch} & \textbf{Event Prog. Ext.} & \textbf{Tokenizer} & \textbf{Tok. Mix.} & \textbf{Chn. Mix.} & \textbf{mAP} & 
             \textbf{Params (M)} & \textbf{Inference (ms)}
             \\
             \hline
             RVT & \checkmark &  & Strided Conv & Multi-axis Attention & MLP & 48.76 
             & 18.54 & 13.51 
             \\
             
             Pool RVT & \checkmark & & Strided Conv & AvgPooling & MLP & 48.54 
             & & 12.03 \\

             MLP RVT & \checkmark &  & Strided Conv & MLP Mixers & MLP & 48.13 
             & 15.90 & \textbf{11.70} \\
             

             EMF Local & \checkmark &  & Root Module & Multi-axis Attention & MLP & 47.43 
             & 17.60 & 15.38 
             \\

             EMF Simple & & & Root Module & RepMixer & ConvFFN & 49.11 
             & \textbf{14.67} & 12.13 \\
             \rowcolor{lgray}
             EMF & & \checkmark & Root Module & RepMixer & ConvFFN & \textbf{50.53} 
             & 14.92 & 12.88 \\

             \hline
        \end{tabular}
    }
\end{table*}

\subsection{Progressive Fine-Tuning}

Progressive fine-tuning demonstrate that performance improvement is possible when a large amount of data is available. In progressive fine-tuning, the network is first trained on a general dataset and then fine-tuned on a target dataset, on which it is finally tested. It is important to note that only the train set is used for training as well as fine-tuning. Tab. \ref{tab:fine_tune} shows detailed results of progressive fine-tuning on Gen1 and 1Mpx datasets. Compared to ERGO \cite{zubic2023chaos} and RVT \cite{gehrig2023recurrent} our proposed \textit{EMF} performs significantly better on 1Mpx and similar to RVT \cite{gehrig2023recurrent} on Gen1 dataset. This proves that our proposed \textit{EMF} scales better with the abundance of data, compared to state-of-the-art methods.

\subsection{Qualitative Comparison}
Fig. \ref{fig:qual_res} shows qualitative comparison of our proposed \textit{EMF} and RVT \cite{gehrig2023recurrent} models. The comparison contains $3$ samples from each Gen1 and 1Mpx datasets. The GT row shows the ground-truths for reference. Samples with only car (cyan) and pedestrian (orange) labels are shown for ease in comparison. It is evident that, our proposed \textit{EMF} model can detect the objects even when RVT misses.

\section{Ablation Study}

We perform ablation study on use of patching and event progression extractor as well as choice of tokenizer, token mixer and channel mixer, to find their contribution towards performance and efficiency metrics. For this purpose, we use the validation set of Gen1 dataset \cite{de2020large}. Tab. \ref{tab:ablation} shows detailed results of this study. We take RVT \cite{gehrig2023recurrent} as a baseline for this experiment. It divides the input into patches to apply multi-axis attention as token mixer while using MLPs as channel mixer. In Pool RVT, we replace multi-axis attention with a simple 2D pooling operation, following the idea of \cite{yu2022metaformer}. This change achieves a slightly poor $mAP$ but significant reduction in inference time, i.e., $1.5\,ms$. In \textit{EMF simple} we do not split the input into patches, and use the \textit{Root Module} as tokenizer, RepMixer as token mixer and ConvFFN as channel mixer. This arrangement achieves a boost of $0.35$ in $mAP$ with a $10\%$ reduction in inference time. We empirically discovered that allowing information to flow between time-bins at an early stage helps the network to better grasp key temporal-features embedded in the channel dimension. To this extent, we employ EPE module on raw event volume before passing it to \textit{EMF Stages}. This simple change results in a significant improvement in $mAP$. We observe an improvement of $1.42$ in $mAP$ with a slight increase in inference time, compared to \textit{EMF simple}. When compared to our baseline, \textit{EMF} achieves an improvement of $1.77$ in $mAP$ with a notable reduction in inference time.
\section{Conclusion}

This paper presents a novel event-based object detection backbone, \textit{EMF}, as an efficient alternative to the state-of-the-art RVT-based backbones \cite{gehrig2023recurrent}. The proposed architecture employs event-tailored feature extractor, replaces computationally demanding modules with convolution-based alternatives, removes patching to improve local features and uses train-time over-parameterization to achieve higher efficiency and state-of-the-art performance. Extensive experiments are performed to evaluate the proposed \textit{EMF} on well-established event-based object detection benchmarks, i.e., Gen1 and 1Mpx datasets. The proposed model achieves state-of-the-art performance on the Gen1 dataset \cite{de2020large} with a significant reduction in inference time. Also, the proposed \textit{EMF} outperforms the most efficient event-based object detector in performance and inference time, to become the fastest DNN-based architecture in the domain. Cross-dataset evaluations and progressive fine-tuning experiments prove that the proposed model achieves superior performance on unseen data and scales better with abundance of data compared to state-of-the-art models.
{
    \small
    \bibliographystyle{ieeenat_fullname}
    \bibliography{main}
}

\end{document}